\documentclass{article}
\usepackage[utf8]{inputenc}
\usepackage{authblk}
\usepackage{setspace}
\usepackage[margin=1.25in]{geometry}

\usepackage{amsmath,amssymb,amsfonts}
\usepackage{algorithmic}
\usepackage{graphicx}
\usepackage{textcomp}
\usepackage{xcolor}
\usepackage{booktabs}
\usepackage{subcaption}
\usepackage{hyperref}
\usepackage{balance}
\usepackage{svg}
\usepackage{soul}
\usepackage[ruled,vlined]{algorithm2e}
\usepackage{wrapfig}

\usepackage[style=ieee, 
citestyle=numeric-comp,
sorting=none]{biblatex}
\title{On Board Volcanic Eruption Detection through CNNs and Satellite Multispectral Imagery}

\author[1*$\dag$]{Maria Pia Del Rosso}
\author[1$\dag$]{Alessandro Sebastianelli}
\author[2,3$\dag$]{Dario Spiller}
\author[2$\dag$]{Pierre Philippe Mathieu}
\author[1$\dag$]{Silvia Liberata Ullo}

\affil[1]{Department of Engineering, University of Sannio, Benevento, Italy.}
\affil[2]{$\Phi$-lab, European Space Agency, Frascati, Italy.}
\affil[3]{Italian Space Agency, Rome, Italy.}
\affil[*]{Corresponding author. Email: mariapia.delrosso@unisannio.it}
\affil[$\dag$]{These authors contributed equally to this work.}

\date{}

\begin{document}

\maketitle

\begin{abstract}
In recent years, the growth of Machine Learning (ML) algorithms has raised the number of studies including their applicability in a variety of different scenarios. Among all, one of the hardest ones is the aerospace, due to its peculiar physical requirements. In this context,  a feasibility study and a first prototype for an Artificial Intelligence (AI) model to be deployed on board satellites are presented in this work. As a case study,  the detection of volcanic eruptions has been investigated as a method to swiftly produce alerts and allow immediate interventions. Two Convolutional Neural Networks (CNNs) have been proposed and designed,  showing how to efficiently implement them  for identifying the eruptions and at the same time adapting their complexity in order to fit on board requirements.
\end{abstract}

\section{Introduction}
Nowadays, Remote Sensing (RS) is seeing its maximum expanse in terms of applicability and use cases, because of the extremely large availability of remote-sensed images, mostly satellite-based, allowing many scientists  from different research fields to approach RS applications \cite{karthikeyan2020review}\cite{sepuru2018appraisal}\cite{chong2017review}\cite{ding2021survey}. A notable example is the increasing use of this type of  data in Earth science and geological fields for monitoring parameters which by their nature are or may become difficult to measure, or that need a lot of time and efforts to be recorded with classical instruments.

Although most of the data processing is usually carried out on ground, there have been in the last years some attempts on bringing  the computation effort, or at least a part of it,  on board the satellites \cite{Hyperscout1}. The ultimate frontier of satellite RS is represented by the implementation of AI algorithms on board the satellites for scene classification, cloud masking, and hazard detection, which has seen  the European Space Agency (ESA) as a pioneer in moving the first steps with the $\phi$-sat 1 satellite, launched on 3 September 2020 \cite{Hyperscout2_spie}\cite{phisat}. With this mission, it has been shown how AI models can help recognizing too cloudy images thus avoiding to download them toward the Ground Stations and therefore reducing the data transmission load \cite{cloudscout}.

The aim of this study is to investigate the possibility of using satellite images to monitor  hazardous events, specifically volcanic eruption, by means of AI  techniques and on board computing resources \cite{book}. The results achieved for the classification of volcanic eruptions could be suitable for future autonomous satellites such as the next ones from the $\phi$-sat program.

In the literature, many researchers made use of satellite images, and Synthetic Aperture Radar (SAR) data in particular, to monitor ground movements in proximity of a volcano's crater just before the eruption. Yet, in recent years, many scientists have switched from classical to AI techniques as for example in \cite{titos2018deep}, where the authors investigated the use of two different Deep Neural Networks (DNNs) together with a combined feature vector of linear prediction coefficients and statistical properties, for seismic events classification purposes. Similar approaches can be found using both real \cite{anantrasirichai2018detecting} and simulated \cite{anantrasirichai2019deep} \cite{sun2020automatic} Interferometric SAR (InSAR) data.

In our work  Sentinel-2 and Landsat-7 optical data have been considered, and based on the current state of the art, the main contributions  are the following:
\begin{enumerate}
    \item  the on board detection of volcanic eruptions by  CNN approaches has never been taken into consideration so far in the literature, to the best of the authors' knowledge.
    \item the proposed CNN  is discussed with regard to the constraints imposed by the on board implementation, which means that the network must be optimized and modified in order to be consistent with target hardware architectures.
    \item the performances of the CNN deployed on the target hardware are analyzed and discussed after the execution of experimental tests.
\end{enumerate}

The rest of the paper is organized as follows. Sec. \ref{sec:datatset} deals with the description of the volcanic eruptions dataset, while Sec. \ref{sec:model} presents the CNN models explored in this study. In Sec. \ref{sec:on_board} the on board implementation of the proposed detection approach is discussed. Results and discussion are presented in Sec. \ref{sec:res}. Conclusions are given at the end.

\section{Dataset}
\label{sec:datatset}
This work focuses on volcanic eruptions by using remote sensing data, hence the first necessary step consists in building the right dataset. 

Since no ready-to-use data were found to perfectly fit the specific task of this work, a specific dataset was built by using an online catalog of volcanic events including geolocalization information \cite{Volcanoes_dataset}. Satellite images acquired for the place and the date of interest were collected and labeled using the open-access Python tool presented in \cite{sebastianelli2020automatic}.

\subsection{The volcanic eruptions catalog}
The dataset has been created by selecting the most recent volcanic eruptions reported in the  Volcanoes of the World (VOTW) catalog by the Global Volcanism Program of the Smithsonian Institution. This is a catalog of Holocene and Pleistocene volcanoes and eruptions from the past 10,000 years \cite{Volcanoes_dataset} up to today. An example of information available in the catalog is reported in Table \ref{tab:sample_of_eruption_catalog}. For the purpose of the dataset creation, the only useful information are the starting date of the eruption, the geographic coordinates and the volcano name, so these information were extracted and stored apart.

\begin{table}[!ht]
    \centering
    \resizebox{\columnwidth}{!}{%
    \begin{tabular}{cccc}
        \toprule
        Eruption Start Time & Volcano name & Latitude (deg) & Longitude (deg)\\
        \midrule
    	2019-06-26 & Ulawun & -5.050 & 151.330\\
    	2019-06-24 & Ubinas & -16.355 & 151.330\\
    	2019-06-22 & Raikoke & 48.292 & 153.250\\
    	2019-06-11 & Piton de la Fournaise& -21.244 & 55.708\\
    	2019-06-01 & Great Sitkin & 52.076 & -176.130\\
    	\bottomrule
    \end{tabular}
    }
    \caption{Sample from the  Volcanoes of the World (VOTW) \cite{Volcanoes_dataset}.}
    \label{tab:sample_of_eruption_catalog}
\end{table}

The images used to create the dataset have been collected using Landsat 7 and Sentinel 2 products, accessed with Google Earth Engine \cite{gorelick2017google}. Specifically, Landsat 7 images have been downloaded considering the period 1999 - 2015, whereas Sentinel 2 images are related to the period 2015 - 2019. 

For the Sentinel 2 data, the level 1-C was selected, comprising 13 spectral bands representing TOA (Top Of Atmosphere) reflectance scaled by 10000 and 3 QA bands, including a bitmask band with cloud mask information. The only product of Landsat 7 available in the Google Earth Engine Catalog is level 2.

Even though the technique described in this work is not limited to the application on these two satellite products, the authors have limited this research to Landsat 7 and Sentinel 2 as they cover the entire period of interest. However, the same approach can be extended to other remote sensing optical products presenting the same bands required for this study, e.g. blue, green, red, and the two SWIR (short-wave infrared) bands located approximately at 1650 nm and 2200 nm. The SWIR bands have been considered in order to better locate and inspect the volcanic eruptions. Indeed, volcanic eruptions can be easily located in RGB wavelengths when they are captured by the satellite camera during the eruptive event, which however is not always the case. After the eruption, the initially red lava gets darker and darker, even though its temperature remains very high. In order to highlight high temperature soil, temperature sensitive bands like the infrared ones have been included.

It is worthy to highlight that there are some differences between the Sentinel-2 and Landsat-7 products, both in terms of spatial resolution and wavelength/bandwidth of the bands of interest, as shown in Table \ref{tab:landsat7_bands} and Table \ref{tab:s2_bands}. These differences are addressed in the next sections.

\begin{table}[!ht]
    \centering
    \resizebox{\columnwidth}{!}{%
    \begin{tabular}{ccccc}
        \toprule
        Band name & Description & Wavelength (nm) & Bandwidth (nm)	& Spatial resolution (m)\\
        \midrule
    	B1 & Blue & 485 & 70 & 30\\
    	B2 & Green & 560 & 80 &  30\\
    	B3 & Red &  660 & 70 & 30\\
    	B5 & SWIR 1 & 1650 & 200 &30\\
    	B7 & SWIR 2 & 2220 & 260 & 30\\
    	\bottomrule
    \end{tabular}
    }
    \caption{Description of Landsat 7 bands.}
    \label{tab:landsat7_bands}
\end{table}

\begin{table}[!ht]
    \centering
    \resizebox{\columnwidth}{!}{%
    \begin{tabular}{ccccc}
        \toprule
        Band name & Description & Wavelength (nm) & Bandwidth (nm) & Spatial resolution (m)\\
        \midrule
    	B2 & Blue & 496.6  (S2A) / 492.1  (S2B) & 66 &10\\
    	B3 & Green & 560 (S2A) / 559 (S2B) & 36 &10\\
    	B4 & Red &  664.5 (S2A) / 665 (S2B) & 31 &10\\
    	B11 & SWIR 1 & 1613.7 (S2A) / 1610.4 (S2B) & 91 (S2A)/ 94 (S2B)& 20\\
    	B12 & SWIR 2 & 2202.4 (S2A) / 2185.7 (S2B) & 175 (S2A)/ 185 (S2B)& 20\\
    	\bottomrule
    \end{tabular}
    }
    \caption{Description of Sentinel 2 bands.}
    \label{tab:s2_bands}
\end{table}

\subsection{Data Preparation and Manipulation}
Satellite data have been downloaded with the above cited tool \cite{sebastianelli2020automatic} that allows to automatically download small patches of images. For this work,  the downloaded patches covered an overall area of 7.5 km$^2$.

After downloading the data, some pre-processing procedures have been applied. Firstly, the images have been resized to $512\times512$ pixels using the Bicubic Interpolation method of Python OpenCV. This procedure mitigates the difference of spatial resolution between Sentinel-2 and Landsat-7. Secondly, infrared bands are combined with RGB bands, in order to visually highlight the color of the volcanic lava, regardless of its color (which is typically red during the eruption and darker a few hours later). Finally, in the experimental phase the proposed algorithm has been adapted and deployed on a Raspberry PI board with a PI camera \cite{raspberrypi}. Since the PI camera only acquires  RGB data, the bands' combination has become necessary to simulate RGB-like images. The bands' combination for highlighting IR spectral answers is given by the equation \cite{volcanoesSentinelhub}:

\begin{subequations}
	\begin{align}
		\textrm{RED}   &= \alpha_1 \cdot B4 + \max(0, \textrm{SWIR2}-0.1)\\
		\textrm{GREEN} &= \alpha_2 \cdot B3 + \max(0, \textrm{SWIR1}-0.1)\\
		\textrm{BLUE}  &= \alpha_3 \cdot B2
	\end{align}
	\label{eqn:band_marging_formula}
\end{subequations}

\begin{figure}[!ht]
	\centering
	\includegraphics[width=\columnwidth]{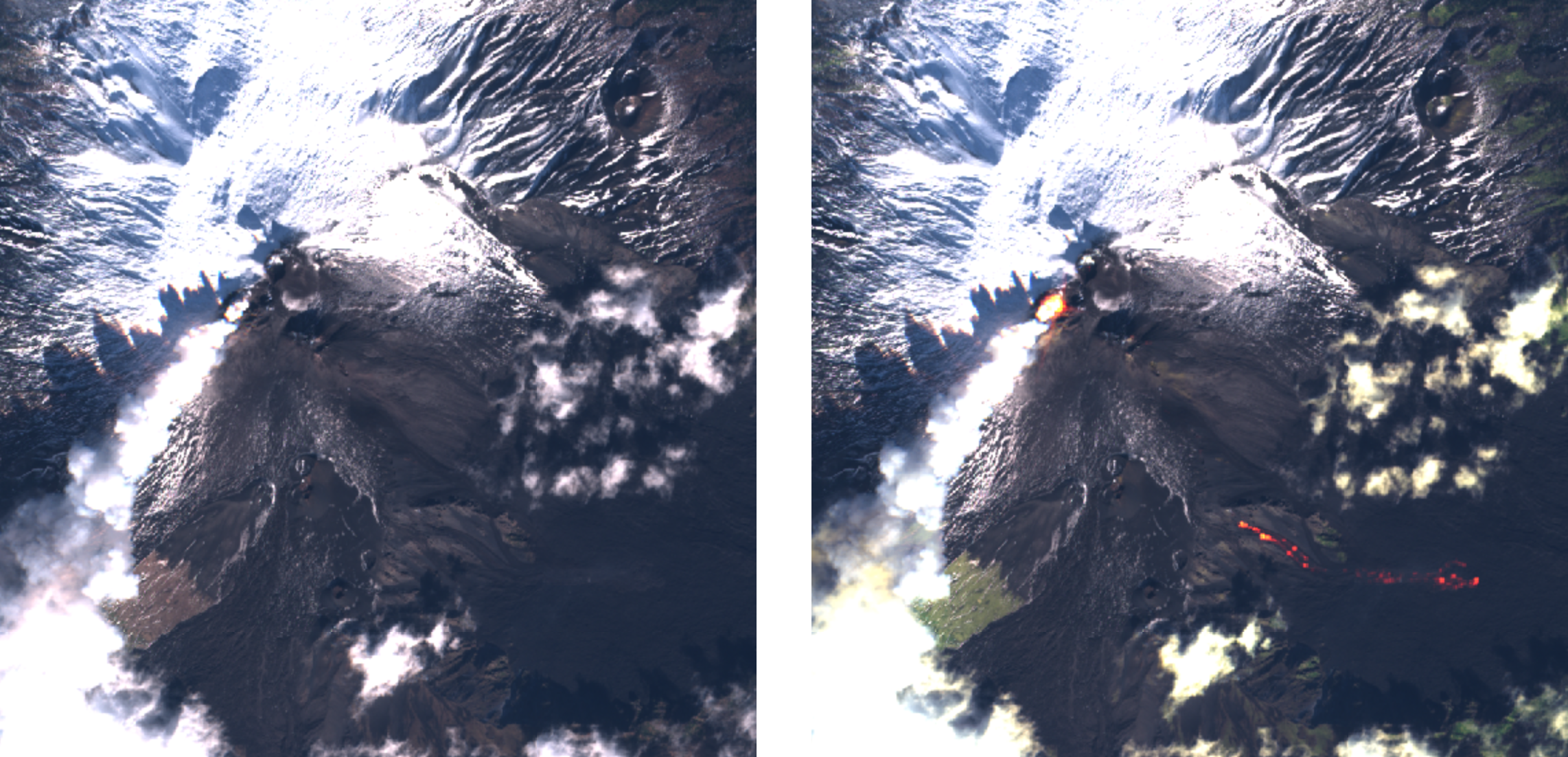}
	\caption{True RGB color image (left) and IR highlighted image (right).}
	\label{fig:true_vs_ir}
\end{figure}

Namely, red and green bands are mixed with SWIR1 and SWIR2 bands, in order to enhance the pixels with high temperature. The multiplicative factor $\alpha_x$ is used to adjust the scale of the image and it is set to $2.5$. In practice, the infrared bands change the red and green bands, so that the heat information is highlighted and visible to the human eye. In this way, it was possible to create a quantitatively correct dataset since, during the labeling, the eruptions were easily distinguishable from non-eruptions. In Figure \ref{fig:true_vs_ir} the difference between a simple RGB image and the  one highlighting IR is shown.

\subsection{Dataset expansion}
\begin{figure}[!ht]
	\centering
	\includegraphics[width=\columnwidth]{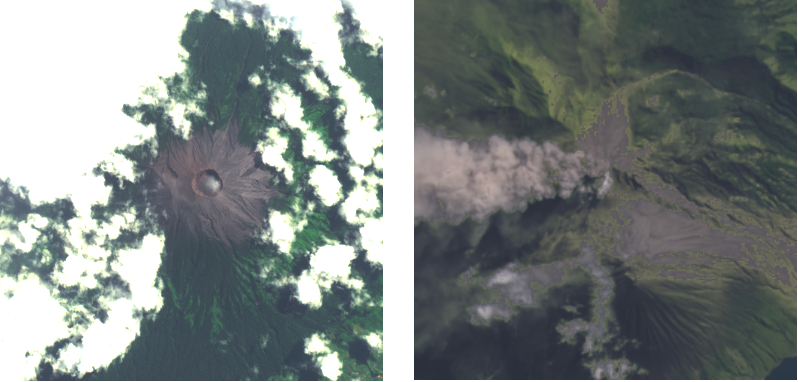}
	\caption{Volcano surrounded by clouds (left) vs Eruption smoke (right)}
	\label{fig:smoke_vs_cloud}
\end{figure}

Since the task addressed in this paper is a typical binary classification problem,  images have been downloaded in order to fill both the eruption and the no-eruption classes. To have a high variability and to reach better results the no-eruption images have been downloaded by focusing on five sub-classes: $1)$ non-erupting volcanoes, $2)$ cities, $3)$ mountains, $4)$ cloudy images and $5)$ completely random images. The presence of cloudy images is really important, in order to make the CNN learn to distinguish between eruption smoke and clouds. An example of comparison is  shown in Figure \ref{fig:smoke_vs_cloud}.

\begin{figure}[!ht]
    \centering
    \includegraphics[width=\columnwidth]{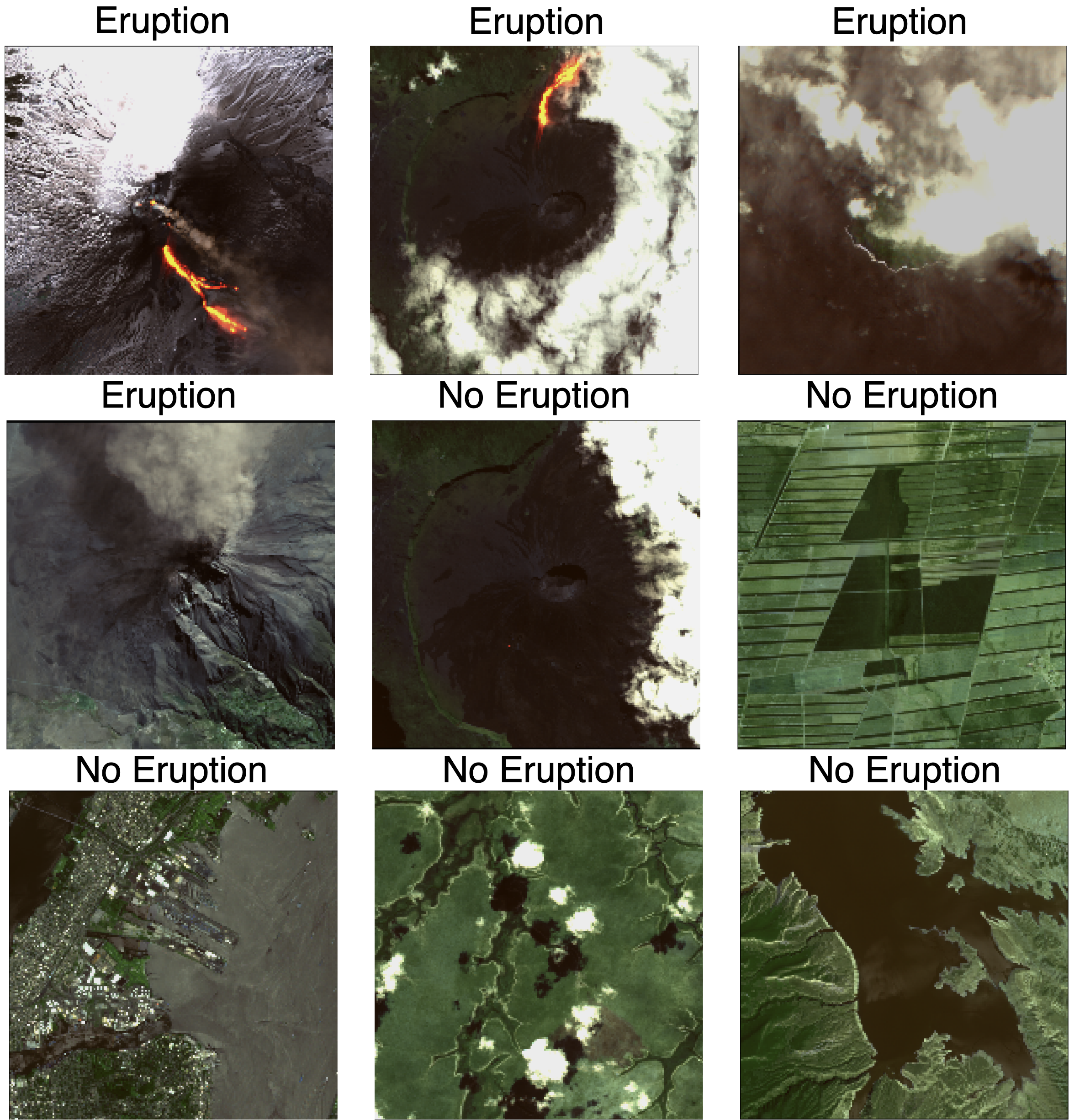}
    \caption{A set of 9 images from the downloaded and pre-processed dataset.}
    \label{fig:dataset_example}
\end{figure}

The same pre-processing step has been applied to the new data, since in the deep learning context a homogeneous dataset is preferable for reducing the sensitivity of the model  to variations in the distribution of the input data \cite{lecun2015deep,goodfellow2016deep}. The final dataset contains 260 images for the class eruption and 1500 for the class non-eruption. Due to the type of event analyzed, the dataset appears to be unbalanced, as an acquisition with an eruption is a rare event. The problem of the imbalanced dataset is addressed in the next sections. Some samples from the dataset are shown in Figure \ref{fig:dataset_example}.

\begin{figure}[!ht]
    \centering
    \includegraphics[width=1\columnwidth]{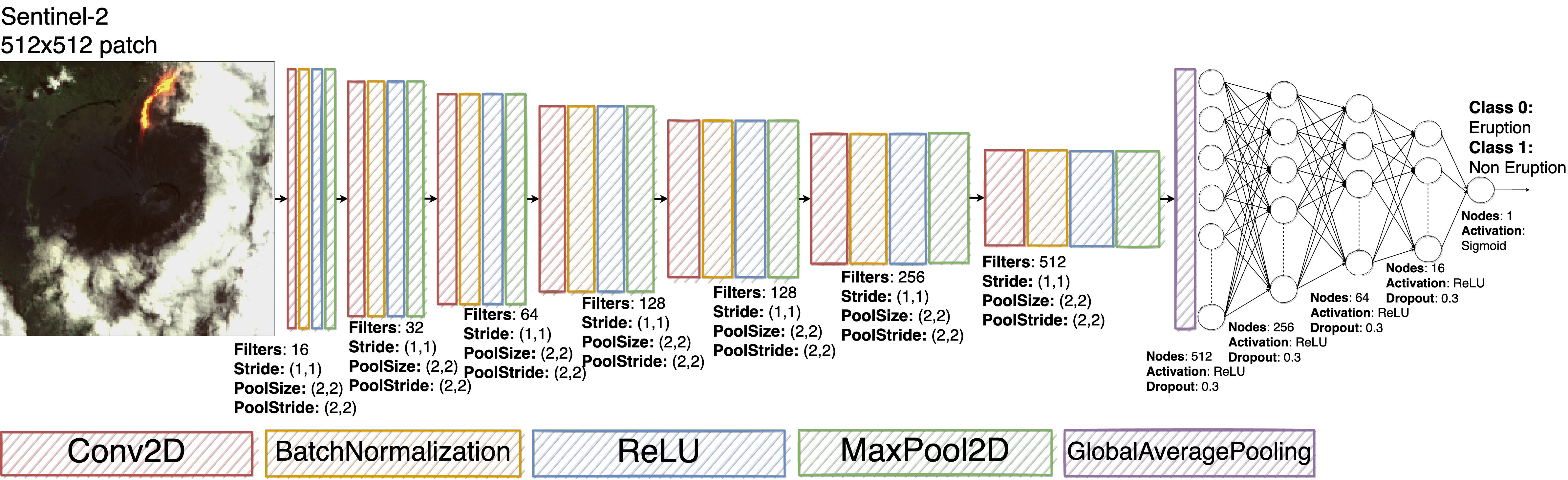}
    \caption{Network architecture}
    \label{fig:model_big}
\end{figure}
\section{Proposed Model}
\label{sec:model}

The detection task has been addressed by implementing a binary classifier where the first class is assigned to images with eruptions and the second one addresses all the other scenarios. The overall CNN architecture is shown in Figure \ref{fig:model_big}. The proposed CNN can be divided in two sub-networks: the first convolutional network is responsible for the features extraction and the second fully connected network is responsible for the classification task \cite{lecun2015deep, goodfellow2016deep, kim2017convolutional}.

The first sub-network consists of seven convolutional layers, each one followed by a batch normalization layer, a ReLU activation function and a max pooling layer. Each convolutional layer has a stride value equal to (1,1) and an increasing number of filters, from 16 to 512. Each max pooling layer (with size (2,2) for both kernel and stride) halves the feature map dimension. The second sub-network consists of five fully-connected layers,  where each layer is followed by a ReLU activation function and a dropout layer. In this case the number of elements of each layer decreases. In the proposed architecture the two sub-networks are connected with a global average pooling layer that, compared to a flatten layer, drastically reduces the number of trainable parameters, speeding up the training process.

\subsection{Image Loader and Data Balancing}
Given the nature of the analyzed hazard, the dataset results unbalanced. An unbalanced dataset, with a number of examples of one class much greater than the other, will lead the model to recognize only the dominant class. To solve this issue, an external function called Image Loader from the Phi-Lab \textit{ai4eo.preprocessing} library has been used \cite{ai4eogithub}.

This library allows the user to define a much more efficient image loader than the already existing Keras version. Furthermore, it is possible to implement a data augmentator that allows the user to define further transformations. The most powerful feature of this library is the one related to the balancing of the dataset through the oversampling technique. In particular, each class is weighted independently using a value depending on the number of samples of the class.

\subsection{Training}
During the training phase, for each epoch the error between the real output and the prediction is calculated both on the training and on the validation datasets. The metric used for the error has been the accuracy and it is worth to underline that this metric works precisely only if there is an equal number of samples belonging to both classes. The model has been trained for 100 epochs, using the Adam optimizer and the binary cross-entropy as loss function.

The training dataset is composed of 1215 examples, among which 334 are with eruptions and 818 are without eruptions. The validation dataset contains 75 eruption samples and 94 no eruption images. Both datasets have been subjected to the data augmentation and to the addition of white Gaussian noise to each channel to increase the robustness of the model and to solve the spectral diversity between Sentinel-2 and Landsat-7.

\begin{figure}[!ht]
    \centering
    \includegraphics[width=\columnwidth]{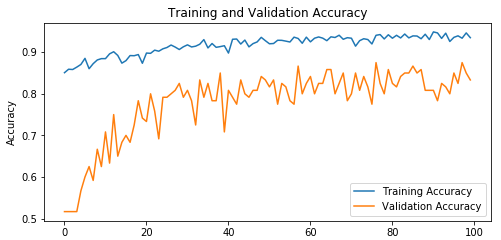}
    \caption{Training and validation accuracy for the proposed model}
    \label{fig:acc_loss}
\end{figure}

The model was trained on the Google Colaboratory platform, where each user can count on: $1)$ a GPU Tesla K80, having 2496 CUDA cores, compute 3.7, 12G GDDR5  VRAM, $2)$ a CPU single core hyper threaded i.e (1 core, 2 threads) Xeon Processors @2.3 Ghz (No Turbo Boost), $3)$ 45MB Cache, $4)$ 12.6 GB of available RAM and $5)$ 320 GB of available disk. With such an architecture, each training epoch required about 370 seconds. The trends of the accuracy on training and validation dataset are shown in Figure \ref{fig:acc_loss}.

\subsection{Model Pruning}
Since the final goal is to realize a model to be uploaded on an on board system, its optimization is necessary  in terms of network complexity, number of parameters and inference execution time. The choice of using a small chip led to limitations of executing the specific classification model, due to the chip's limited elaboration power, thus it was necessary to derive a proper model.

Hence, starting from the first, feasible network, a second and smaller network was built by pruning the former one. The convolutional sub-network has been reduced by removing three layers and the fully connected sub-network has been drastically reduced to only two layers. The choice on the number of layers to remove has been a trade off between the number of network parameters and its capability of extracting useful features. For this reason, the choice relapsed on discarding the last two convolutional layers and the very first one. This choice led to quite good results, revealing a good compromise between memory weight, computational load and performances. The new modified network has shown to be still capable of discriminating or classifying data correctly, with an accuracy of 0.83. 

\begin{figure}[!ht]
    \centering
    \includegraphics[width=1\columnwidth]{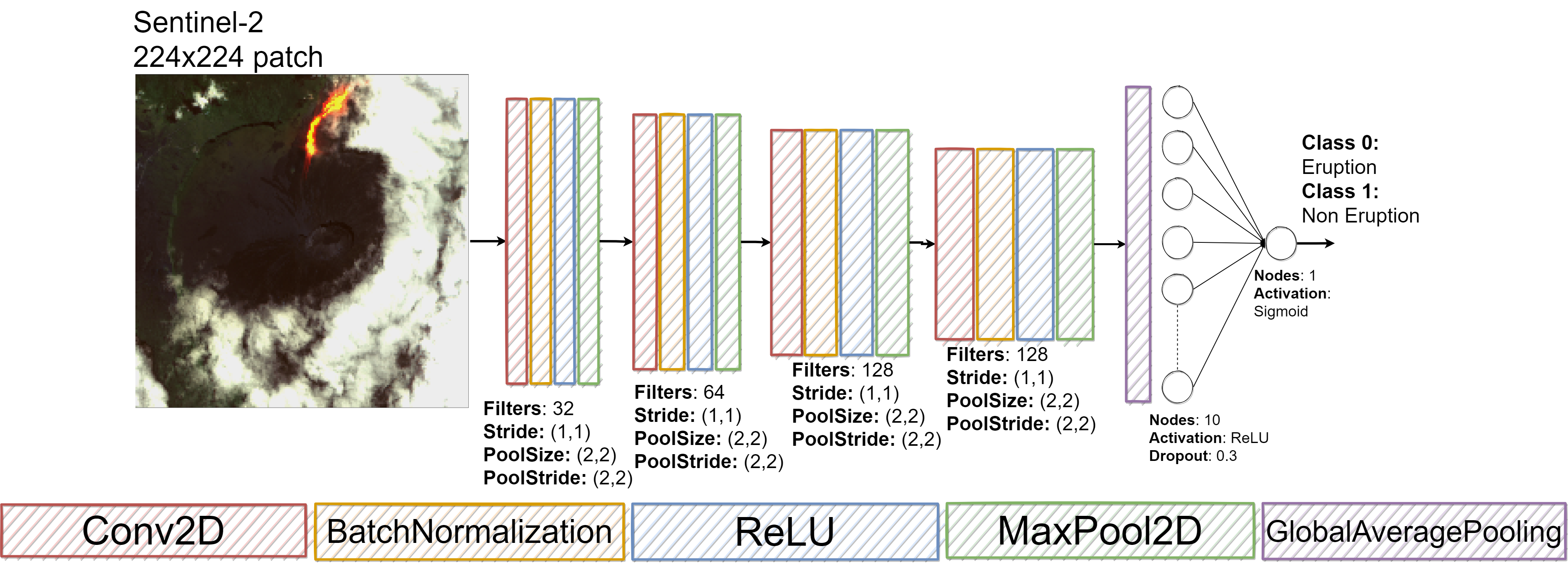}
    \caption{Smaller Network Architecture}
    \label{fig:small_model}
\end{figure}

\begin{figure}[!ht]
    \centering
    \includegraphics[width=1\columnwidth]{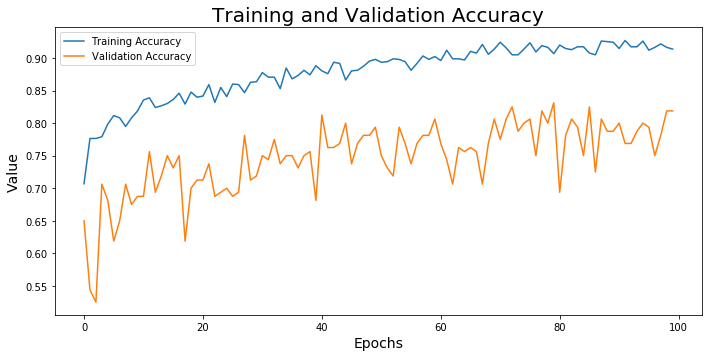}
    \caption{Training and validation accuracy for the proposed reduced model}
    \label{fig:small_acc_loss}
\end{figure}

The smaller model has been trained with the same configuration of the original one. The trends of the training and the validation loss and accuracy functions are shown in Figure \ref{fig:small_acc_loss}.

\section{Going on board, a first prototype}
\label{sec:on_board}

\begin{figure}[!ht]
    \centering
    \includegraphics[width=\columnwidth]{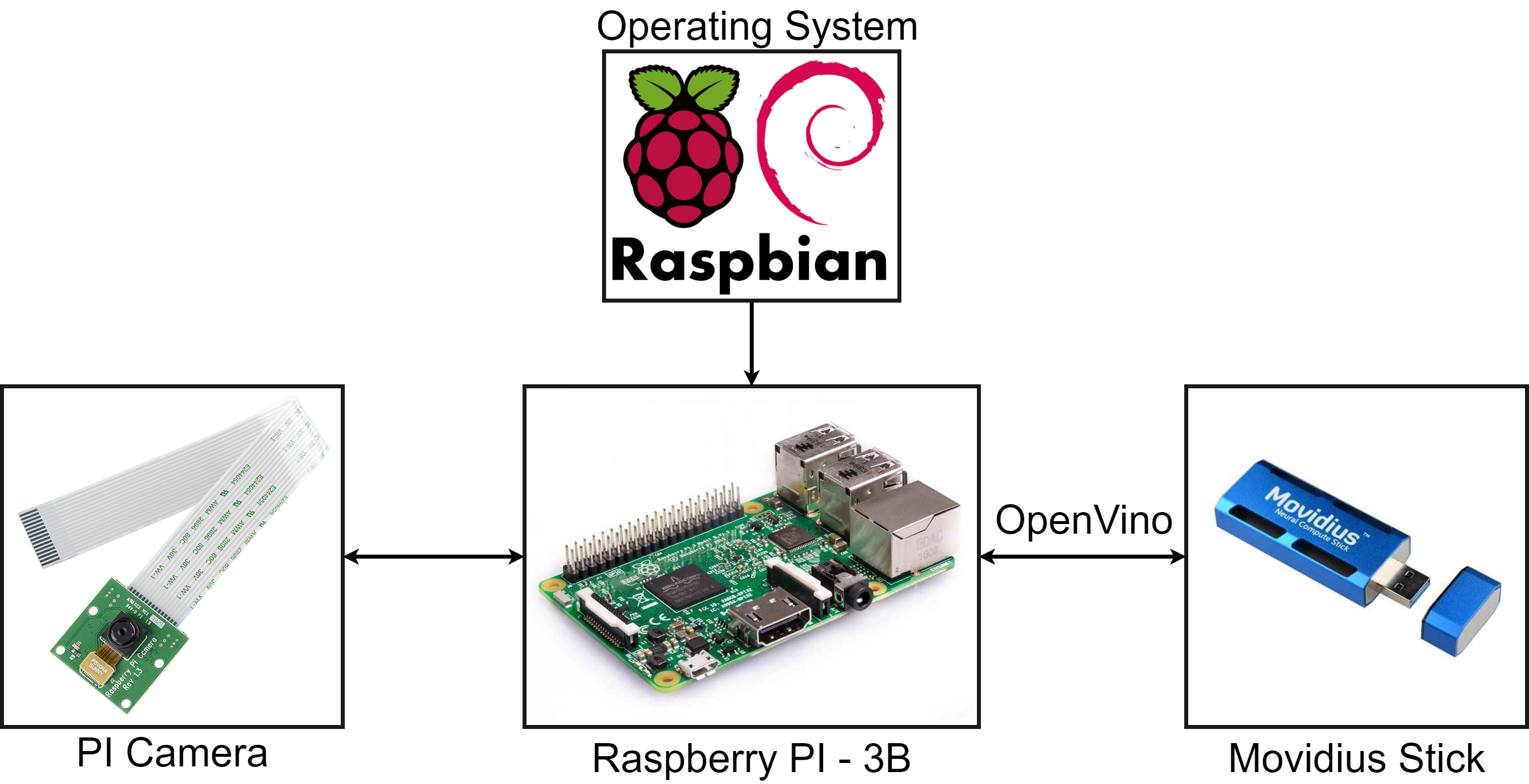}
    \caption{Schematic for the prototype}
    \label{fig:proto}
\end{figure}

The proposed models have been developed for carrying out an  on board volcanic eruption classification. A prototype has been assembled and mounted on a drone; it was mainly composed of a Raspberry PI, a PI camera and a Movidius Stick for Deep Learning purposes \cite{inteldatasheet, raspberrycameradatasheet, raspberrydatasheet}, as shown in Figure \ref{fig:proto}.

The description of the architecture of the drone system is out of the scope of this work since the drone has only been used for simulation purposes, thus its subsystems are not included in the schematic.

The Raspberry PI, with the Raspbian Operating System (OS), is the on board computer that process the images acquired through the PI camera and sends them to the Movidius Stick to run the classification algorithm based on the CNN model. It is also responsible for sending the prediction results to the local PC through the Wi-Fi.

\subsection{Raspberry PI}
The Raspberry adopted for this use case is the Raspberry Pi 3 Model B, the earliest model of the third-generation Raspberry Pi. Table \ref{tab:rasp_spec} shows its main specifications.

\begin{table}[!ht]
    \centering
    \resizebox{1\columnwidth}{!}{%
    \begin{tabular}{cc}
    \toprule
    Component & Specifications\\
    \midrule
    Processor & Quad Core 1.2GHz Broadcom BCM2837 64bit CPU 1GB RAM\\
    Wireless systems & BCM43438 wireless LAN and Bluetooth Low Energy (BLE) on board\\
    Hardware sysetms & 100 Base Ethernet\\
    Hardware connectors & 40-pin extended GPIO\\
    USB Ports & 4 USB 2.0 ports\\
    Video Ports (1) & 4 Pole stereo output and composite video port\\
    Video Ports (2) & Full size HDMI\\
    Camera Port & CSI camera port for connecting a Raspberry Pi camera\\
    Display Port & DSI display port for connecting a Raspberry Pi touchscreen display\\
    External Memory Port & Micro SD port for loading your operating system and storing data\\
    Power Supply Poprt & Upgraded switched Micro USB power source up to 2.5A\\
	\bottomrule
    \end{tabular}
    }
    \caption{Raspberry PI 3 specifications}
    \label{tab:rasp_spec}
\end{table}

\subsection{Camera}
The Raspberry Pi Camera Module v2, is a high quality 8 megapixel Sony IMX219 image sensor custom designed add-on board for Raspberry Pi, featuring a fixed focus lens. It is capable of 3280x2464 pixel static images, and supports 1080p30, 720p60 and 640x480p90 video.

The camera can be plugged using the dedicated socket and CSi interface. The main specifications for PI camera are listed in Table \ref{tab:camera}.

\begin{table}[!ht]
    \centering
    \resizebox{1\columnwidth}{!}{%
    \begin{tabular}{cc}
    \toprule
    Component & Specifications\\
    \midrule
    Focus & Fixed focus lens on board\\
    Resolution & 8 megapixel native resolution sensor\\
    Frame size & 3280 x 2464 pixel static images\\
    Video Support & Supports 1080p30, 720p60 and 640x480p90 video\\
    Physical Dimension & Size 25mm x 23mm x 9mm\\
    Weights & Weight just over 3g\\
	\bottomrule
    \end{tabular}
    }
    \caption{Raspberry Pi RGB camera specifications}
    \label{tab:camera}
\end{table}

\subsection{Movidius Stick}
The Intel Movidius Neural Compute Stick is a small fanless deep learning USB drive designed to learn AI programming. The stick is powered by the low power high performance Movidius Visual Processing Unit. The main specifications are:
\begin{itemize}
    \item Supporting CNN profiling, prototyping and tuning workflow
    \item Real-time on device inference (Cloud connectivity not required)
    \item Features the Movidius Vision Processing Unit with energy-efficient CNN processing
    \item All data and power provided over a single USB type A port
    \item Run multiple devices on the same platform to scale performance
\end{itemize}

\subsection{Implementation on Raspberry and Movidius Stick}
\begin{figure}[!ht]
    \centering
    \includegraphics[width=1\columnwidth]{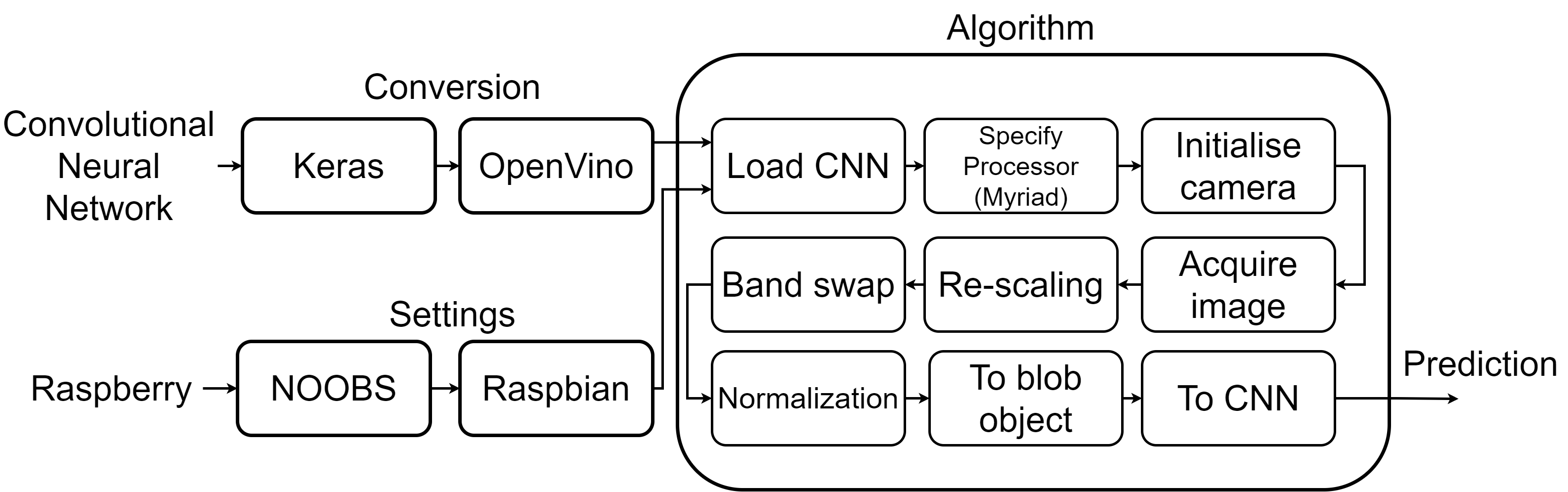}
    \caption{Block diagram for the implementation on Raspberry and Movidius stick}
    \label{fig:script_scheme}
\end{figure}

In order to run the experiments with the Raspeberry PI and the Movidius Stick, two preliminary steps are necessary:\\
$1)$ the CNN must be converted from the original format (e.g. Keras model) to an OpenVino format, using the OpenVino library; $2)$ an appropriate operating system must be install on the Raspberry (e.g. the Rasbian OS through the NOOBS installer). The implementation process is schematized in Figure \ref{fig:script_scheme}.

\begin{figure}[!ht]
    \centering
    \includegraphics[width=1\columnwidth]{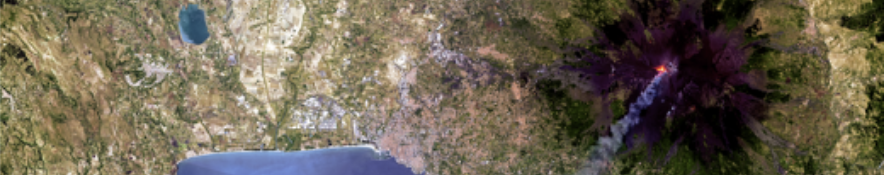}
    \caption{Sentinel-2 printed image for on board model evaluation}
    \label{fig:s2strips}
\end{figure}

\subsubsection{OpenVINO library}
For deep learning, the current Raspberry PI hardware is inherently resource constrained. The Movidius Stick allows faster inference with the deep learning coprocessor that is plugged into the USB socket. In order to transfer the CNN on the Movidius, the network should be optimized, using the OpenVINO Intel library for hardware optimized computer vision.

The OpenVINO toolkit is an Intel Distribution and is extremely simple to use. Indeed, after setting the  target processor, the OpenVINO-optimized OpenCV can handle the overal setup \cite{openvino}.  Based on CNN, the toolkit extends workloads across Intel hardware (including accelerators) and maximizes performance by:

\begin{itemize}
    \item enabling deep learning inference at the edge
    \item supporting heterogeneous execution across computer vision accelerators (e.g. CPU, GPU, Intel Movidius Neural Compute Stick, and FPGA) using a common API
    \item speeding up time to market via a library of functions and pre-optimized kernels
    \item including optimized calls for OpenCV.
\end{itemize}

After implementation on Raspberry and Movidius, the models were tested by acquiring images using a drone flying over a print made with Sentinel-2 data of an erupting volcano, as shown in Figure \ref{fig:s2strips}. The printed image has been processed with the same settings used for the training and validation dataset.

\section{Results and discussion}\label{sec:res}
\subsection{Training on the PC}
\begin{figure}[!ht]
    \centering
    \includegraphics[width=1\columnwidth]{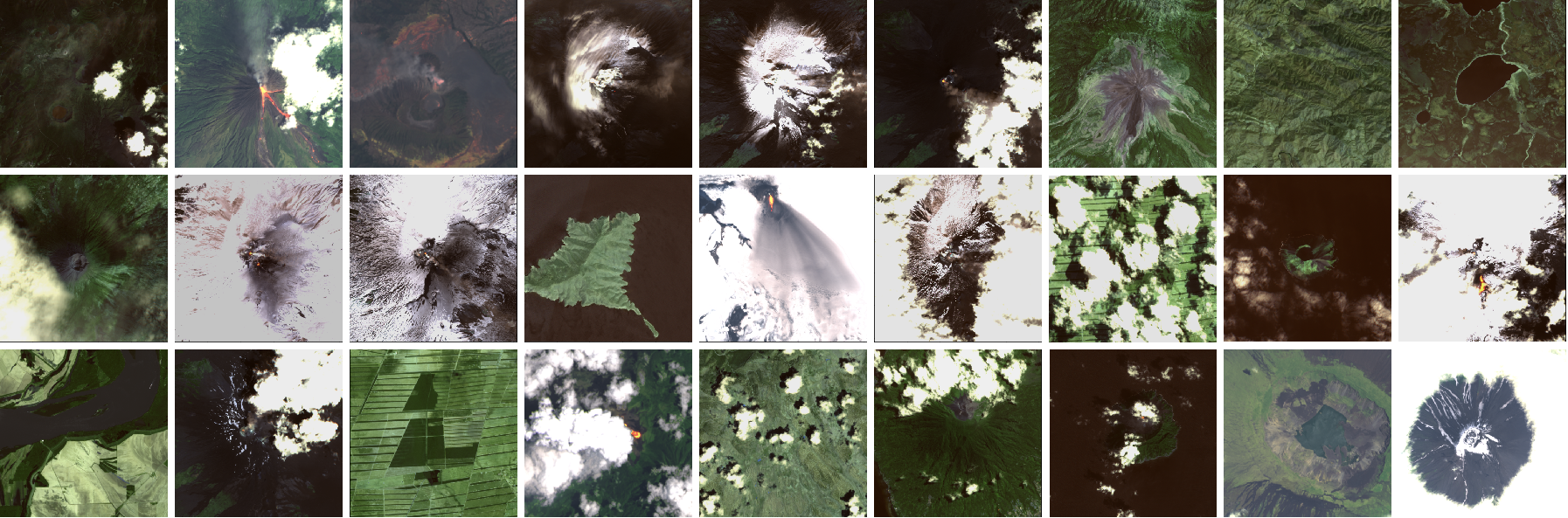}
    \caption{Sample from the testing dataset}
    \label{fig:big_model_tests}
\end{figure}

The model performances have been computed on the testing dataset. In Figure \ref{fig:big_model_tests} 18 samples are shown, while in Table \ref{tab:big_model_tests} the corresponding ground truth and prediction labels are reported. A low value of predicted label indicates a low probability of eruption, so the image is classified as no eruption; a high value of predicted label, indicates a high probability of eruption, so the image is classified as eruption; a value next to 0.5 indicates a random prediction. Since the problem is a binary classification, a threshold value of 0.5 was selected to identify the class based on the prediction: a value lower that 0.5 was rounded to 0 (non-eruption class), while a value higher than the 0.5 was rounded to 1 (eruption class).

\begin{table}[!ht]
    \centering
    \resizebox{\columnwidth}{!}{%
    \begin{tabular}{cccc}
    \toprule
    Column & Row 1 & Row 2 & Row 3\\
    \midrule
    & \begin{tabular}{cc} Ground truth & Predicted\end{tabular} & 
               \begin{tabular}{cc}Ground truth & Predicted\end{tabular} & 
               \begin{tabular}{cc}Ground truth & Predicted\end{tabular}\\
    \midrule    
    1 & \begin{tabular}{cc}1.00 & 0.13 \end{tabular} & \begin{tabular}{cc}0.00 & 0.03 \end{tabular} & \begin{tabular}{cc}0.00 & 0.00 \end{tabular}\\
    2 & \begin{tabular}{cc}1.00 & 0.99 \end{tabular} & \begin{tabular}{cc}1.00 & 0.99 \end{tabular} & \begin{tabular}{cc}1.00 & 0.99 \end{tabular}\\
    3 & \begin{tabular}{cc}1.00 & 0.99 \end{tabular} & \begin{tabular}{cc}1.00 & 0.99 \end{tabular} &  \begin{tabular}{cc}0.00 & 0.00 \end{tabular}\\
    4 & \begin{tabular}{cc}1.00 & 0.95 \end{tabular} & \begin{tabular}{cc}0.00 & 0.00 \end{tabular} & \begin{tabular}{cc}1.00 & 0.99 \end{tabular}\\
    5 & \begin{tabular}{cc}1.00 & 0.98 \end{tabular} & \begin{tabular}{cc}1.00 & 0.99 \end{tabular} & \begin{tabular}{cc}0.00 & 0.00 \end{tabular}\\
    6 & \begin{tabular}{cc}1.00 & 0.99 \end{tabular} & \begin{tabular}{cc}1.00 & 0.99 \end{tabular} & \begin{tabular}{cc}0.00 & 0.02 \end{tabular}\\
    7 & \begin{tabular}{cc}0.00 & 0.00 \end{tabular} & \begin{tabular}{cc}0.00 & 0.00 \end{tabular} & \begin{tabular}{cc}1.00 & 0.99 \end{tabular}\\
    8 & \begin{tabular}{cc}0.00 & 0.00 \end{tabular} & \begin{tabular}{cc}0.00 & 0.26 \end{tabular} & \begin{tabular}{cc}0.00 & 0.88 \end{tabular}\\
    9 & \begin{tabular}{cc}0.00 & 0.00 \end{tabular} & \begin{tabular}{cc}1.00 & 0.99 \end{tabular} & \begin{tabular}{cc}0.00 & 0.01 \end{tabular}\\
	\bottomrule
    \end{tabular}
    }
    \caption{Model results on testing dataset}
    \label{tab:big_model_tests}
\end{table}

The performances have been computed also for the second model, the one obtained after the pruning process. The same samples shown in Figure \ref{fig:big_model_tests} have been used to make the comparison with the original model. Table \ref{tab:small_model_tests} shows the results.

\begin{table}[!ht]
    \centering
    \resizebox{1\columnwidth}{!}{%
    \begin{tabular}{cccc}
    \toprule
    Column & Row 1 & Row 2 & Row 3\\
    \midrule
     & \begin{tabular}{cc} Ground truth & Predicted\end{tabular} 
     & \begin{tabular}{cc}Ground truth & Predicted\end{tabular} 
     & \begin{tabular}{cc}Ground truth & Predicted\end{tabular}\\
    \midrule    
    1 & \begin{tabular}{cc}1.00 & 0.99 \end{tabular} & \begin{tabular}{cc}0.00 & 0.56 \end{tabular} & \begin{tabular}{cc}0.00 & 0.08 \end{tabular}\\
    2 & \begin{tabular}{cc}1.00 & 0.99 \end{tabular} & \begin{tabular}{cc}1.00 & 0.94 \end{tabular} & \begin{tabular}{cc}0.00 & 0.00 \end{tabular}\\
    3 & \begin{tabular}{cc}1.00 & 0.99 \end{tabular} & \begin{tabular}{cc}1.00 & 0.97 \end{tabular} & \begin{tabular}{cc}0.00 & 0.01 \end{tabular}\\
    4 & \begin{tabular}{cc}1.00 & 0.01 \end{tabular} & \begin{tabular}{cc}0.00 & 0.01 \end{tabular} & \begin{tabular}{cc}0.00 & 0.64 \end{tabular}\\
    5 & \begin{tabular}{cc}1.00 & 0.99 \end{tabular} & \begin{tabular}{cc}1.00 & 0.99 \end{tabular} & \begin{tabular}{cc}0.00 & 0.00 \end{tabular}\\
    6 & \begin{tabular}{cc}1.00 & 0.97 \end{tabular} & \begin{tabular}{cc}1.00 & 0.99 \end{tabular} & \begin{tabular}{cc}1.00 & 0.97 \end{tabular}\\
    7 & \begin{tabular}{cc}0.00 & 0.00 \end{tabular} & \begin{tabular}{cc}0.00 & 0.99 \end{tabular} & \begin{tabular}{cc}1.00 & 0.00 \end{tabular}\\
    8 & \begin{tabular}{cc}0.00 & 0.99 \end{tabular} & \begin{tabular}{cc}0.00 & 0.00 \end{tabular} & \begin{tabular}{cc}0.00 & 0.00 \end{tabular}\\
    9 & \begin{tabular}{cc}0.00 & 0.00 \end{tabular} & \begin{tabular}{cc}1.00 & 0.00 \end{tabular} & \begin{tabular}{cc}0.00 & 0.00 \end{tabular}\\
	\bottomrule
    \end{tabular}
    }
    \caption{Small Model results on testing dataset}
    \label{tab:small_model_tests}
\end{table}

\subsection{Results on the Movidius}
The confusion matrix and the architecture computational speed, for the two models, are reported in Table \ref{tab:smallvsbig}, where is evident that the performances in terms of good prediction are slightly changed after the model pruning, but the most important aspect is that the images per second that the model and the hardware can handle is increased from 1 to 7. 

\begin{table}[!ht]
    \centering
    \resizebox{0.7\columnwidth}{!}{%
    \begin{tabular}{ccc}
        \toprule
        Score & Big Model & Small Model\\
        \midrule
        True Positive  & 0.85 & 0.83 \\ 
        True Negative  & 0.85 & 0.83 \\
        False Positive & 0.15 & 0.17 \\
        False Negative & 0.15 & 0.17 \\
        \midrule
        images/second  &    1 &   7  \\
    	\bottomrule
    \end{tabular}
    }
    \caption{Performances comparison considering the big and the small models.}
    \label{tab:smallvsbig}
\end{table}

This is the fundamental result, as an increase in the processing frequency, maintaining good performances, in our specific case, has allowed us to respect the constraints imposed by the dynamics and mission of the drone used in the experimental phase. This result is perfectly transferable to the satellite domain, obviously considering the constraints imposed by these platforms.

The results are promising, demonstrating the correctness of the proposed pipeline and justifying further analysis and investigation.

\section{Conclusions}
This work aimed to present a first workflow on how to develop and implement an AI model to be suitable for being carried on board satellites for Earth Observation. In particular, two detectors of volcanic eruptions have been developed and discussed. 

AI on board is a very challenging field of research which has seen the European Space Agency (ESA) as a pioneer in moving the first steps with the $\phi$-sat 1 satellite launched on 3 September 2020. 
The possibility to process data  on board a satellite can drastically reduce the time between the image acquisition and its analysis, completely deleting the time for downlinking and reducing the total latency. In this way, it is possible to produce fast alerts and interventions when hazardous events are going to happen.

A prototype and a simulation process, keeping a low-cost kind of implementation, were realized. The experiment and the development chain were completed with commercial and ready-to-use hardware components,  and a drone was also employed for simulations and testing. The AI processor had no problem recognizing the eruption in the test printed image. The results are encouraging, showing that even the pruned model can reach a good performance in detecting the eruptions. Further studies will help to understand possible extensions and improvements.

\section*{Acknowledgments}
The research work published in this manuscript has been developed in collaboration with the European Space Agency (ESA) $\Phi$-lab \cite{philab} during the traineeship of Maria Pia Del Rosso and Alessandro Sebastianelli started in  2019. This research is also supported by the ongoing Open Space Innovation Platform (OSIP) project started in June 2020 and titled "Al powered cross-modal adaptation techniques applied to Sentinel-1 and -2 data",  under a joint collaboration between the ESA $\Phi$-Lab and the University of Sannio.   The results achieved for the classification of volcanic eruptions could be suitable for future missions of the $\phi-sat$ program that represents the first experiment carried out by ESA to demonstrate how AI on board the satellites can be used for Earth Observation \cite{phisat}.

\printbibliography

\section*{Short Biography of Authors}
\begin{wrapfigure}{l}{25mm} 
    \includegraphics[width=1in,height=1.25in,clip,keepaspectratio]{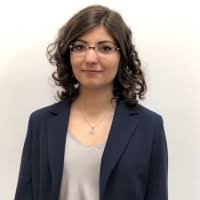}
\end{wrapfigure}\par
\textbf{Maria Pia Del Rosso} graduated  with  laude  in  Electronics Engineering for Automation and Telecommunications  at  the  University  of Sannio  in  October 2019 and she is currently a PhD candidate.  As a master student, she has been a visiting researcher at the Phi-lab in the European Space  Research  Institute  (ESRIN)  of  the  European  Space Agency (ESA), in Frascati. She worked on applying Deep Learning techniques to Remote Sensing Earth Observation data for monitoring geohazard phenomena, such as earthquakes, landslides and volcanic eruptions. She is currently working on the matching between multispectral and radar data applying AI techniques.\par
  
\begin{wrapfigure}{l}{25mm} 
    \includegraphics[width=1in,height=1.25in,clip,keepaspectratio]{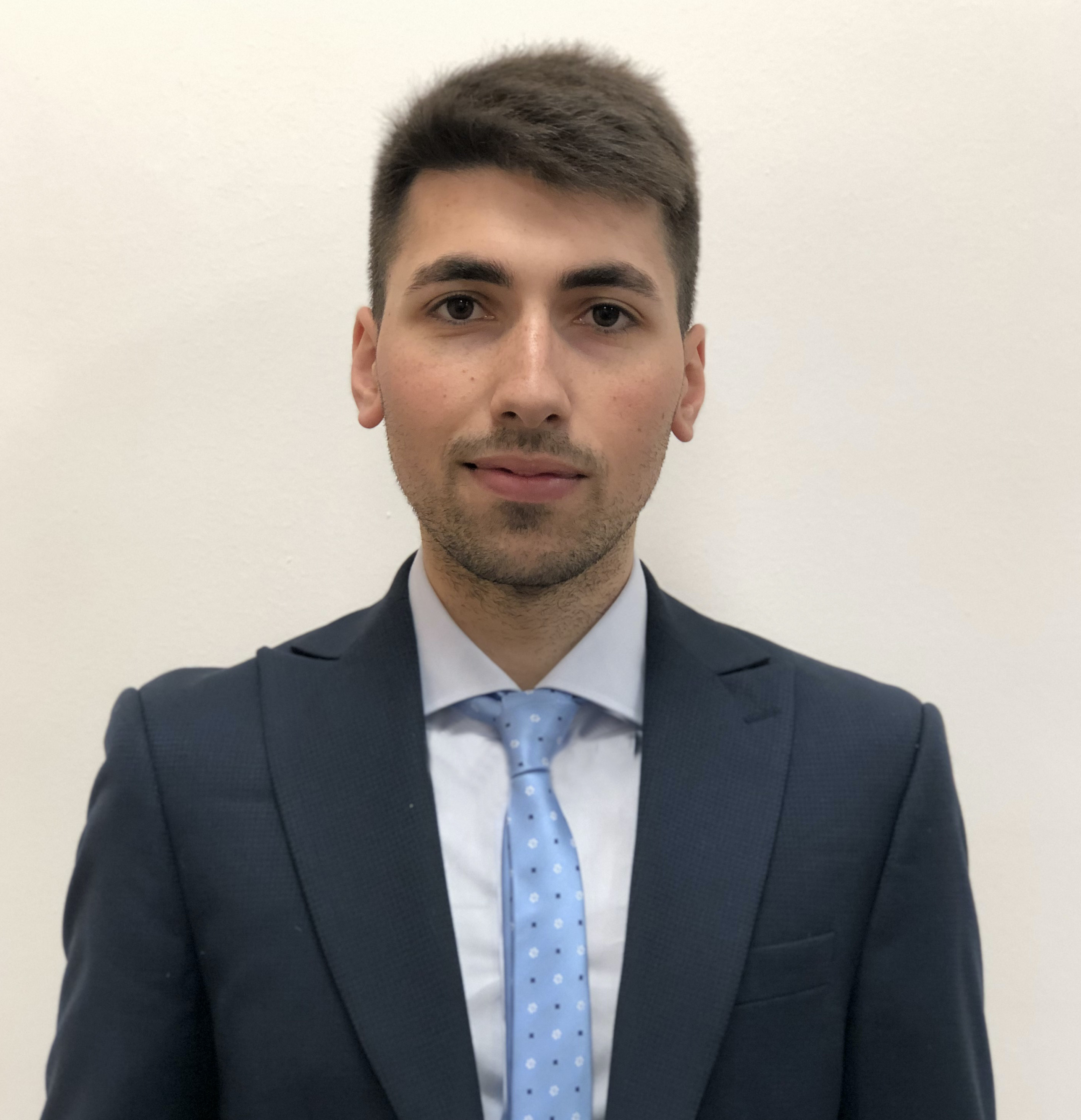}
\end{wrapfigure}\par
\textbf{Alessandro Sebastianelli} \textbf{Alessandro Sebastianelli} graduated  with laude in Electronic Engineering for Automation and Telecommunications at the University  of Sannio in 2019. He is enrolled in the Ph.D. program with University of Sannio, and his research topics mainly focus on Remote Sensing and Satellite data analysis, Artificial  Intelligence  techniques for Earth Observation, and data fusion. He has co-authored several papers to reputed journals and conferences for  the  sector  of  Remote Sensing. Ha has been a visited researcher at Phi-lab in European Space  Research  Institute  (ESRIN)  of  the  European  Space Agency (ESA), in Frascati, and still collaborates with the $\Phi$-lab  on  topics  related  to  deep  learning  applied  to  geohazard assessment,  especially  for  landslides,  volcanoes,  earthquakes phenomena. He has won an ESA OSIP proposal in August 2020 presented with his Ph.D. Supervisor, Prof. Silvia L. Ullo.\par

\begin{wrapfigure}{l}{25mm} 
    \includegraphics[width=1in,height=1.25in,clip,keepaspectratio]{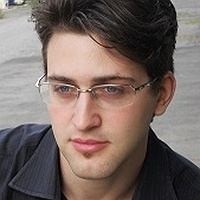}
\end{wrapfigure}\par
\textbf{Dario Spiller} is a PostDoc research fellow working for a joint research project of the Italian space agency (ASI) and the European space agency (ESA). He is an aerospace engineer with a Ph.D. in optimal control based on meta-heuristic optimization applied to space problems related to attitude and orbital maneuvers. Currently, his research is focusing on classification and regression problems applied to remote sensing test cases and solved with machine learning algorithms. He is mainly working on hyperspectral remote sensing and the PRISMA mission with application to wildfire detection and crop type classification.\par

\begin{wrapfigure}{l}{25mm} 
    \includegraphics[width=1in,height=1.25in,clip,keepaspectratio]{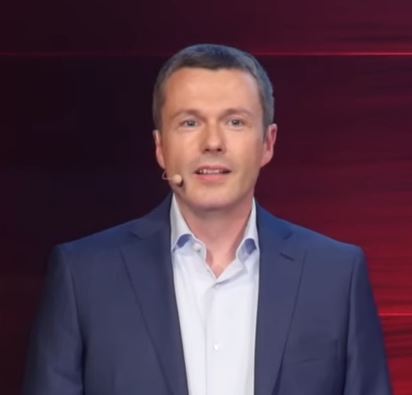}
\end{wrapfigure}\par
\textbf{Pierre Philippe Mathieu} received the M.Sc. degree in engineering from the University of Liege, Liege, Belgium, the Ph.D. degree in climate science from the University of Louvain, Louvain, Belgium, and the Management degree from the University of Reading Business School, Reading, U.K.,Over the last 20 years, he has been working in environmental monitoring and modeling, across disciplines from remote sensing, modeling, up to weather risk management. He is currently a Data Scientist with the European Space Research Institute, European Space Agency, Frascati, Italy, working to foster the use of our Earth observation missions to support science, innovation, and development in partnership with user communities, industry, and businesses. His particular interest lies in addressing global environmental and development issues related to management of food water energy resources and climate change. He is currently the head of $\Phi$-lab explore office of the European Space Agency.\par

\begin{wrapfigure}{l}{25mm} 
    \includegraphics[width=1in,height=1.25in,clip,keepaspectratio]{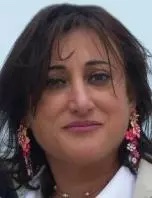}
\end{wrapfigure}\par
\textbf{Silvia Liberata Ullo} IEEE Senior Member, Industry Liaison for IEEE Joint ComSoc/VTS Italy Chapter. National Referent for FIDAPA BPW Italy Science and Technology Task Force. Researcher since 2004 in the Engineering Department of the University of Sannio, Benevento (Italy). Member of the Academic Senate and the PhD Professors’ Board. She is teaching: Signal theory and elaboration, and Telecommunication networks for Electronic Engineering, and Optical and radar remote sensing for the Ph.D. course. Authored 80+ research papers, co-authored many book chapters and served as editor of two books, and many special issues in reputed journals of her research sectors. Main interests: signal processing, remote sensing, satellite data analysis, machine learning and quantum ML, radar systems, sensor networks, and smart grids. Graduated with laude in 1989 in Electronic Engineering,  at the Faculty of Engineering at the Federico  II  University, in  Naples, she pursued     the     M.Sc.     degree     from     the Massachusetts Institute  of Technology (MIT) Sloan  Business  School  of  Boston,  USA,  in June 1992.  She has worked in the private and public sector from 1992 to 2004, before joining the University of Sannio..\par

\end{document}